\documentclass[runningheads]{llncs}

\usepackage{multirow}
\usepackage{xcolor}
\usepackage[T1]{fontenc}
\usepackage{graphicx}
\usepackage{amsmath,amssymb,amsfonts}
\usepackage{cite}
\usepackage{textcomp}
\usepackage{hyperref}
\usepackage{float}
\usepackage{algorithmic}
\usepackage{booktabs}
\usepackage{subcaption}

\usepackage[margin=3cm]{geometry}

\usepackage{etoolbox}
\begin{document}

\title{Multivariate Time-series Anomaly Detection via \\ Dynamic Model Pool \& Ensembling}
\titlerunning{Multivariate Time-series Anomaly Detection via Dynamic Model Pool \& Ensembling}

\author{
Wei Hu,
Zewei Yu,
and Jianqiu Xu
}

\authorrunning{W. Hu et al.}

\institute{
College of Computer Science and Technology \\
Nanjing University of Aeronautics and Astronautics \\
Nanjing 211106, China \\
\email{\{hwei, yuzewei, jianqiu\}@nuaa.edu.cn} \\
Corresponding author: Jianqiu Xu
}

\maketitle

\begin{abstract}
Multivariate time-series (MTS) anomaly detection is critical in domains such as service monitor, IoT, and network security.
While multi-model methods based on selection or ensembling outperform single-model ones, they still face limitations: (i) selection methods rely on a single chosen model and are sensitive to the strategy; (ii) ensembling methods often combine all models or are restricted to univariate data; and (iii) most methods depend on fixed data dimensionality, limiting scalability.
To address these, we propose DMPEAD, a Dynamic Model Pool and Ensembling framework for MTS Anomaly Detection. The framework first (i) constructs a diverse model pool via parameter transfer and diversity metric, then (ii) updates it with a meta-model and similarity-based strategy for adaptive pool expansion, subset selection, and pool merging, finally (iii) ensembles top-ranked models through proxy metric ranking and top-k aggregation in the selected subset, outputting the final anomaly detection result.
Extensive experiments on 8 real-world datasets show that our model outperforms all baselines, demonstrating superior adaptability and scalability.

\end{abstract}


\section{Introduction}
\begin{figure}[htbp]
    \centering
    \includegraphics[width=0.5\textwidth]{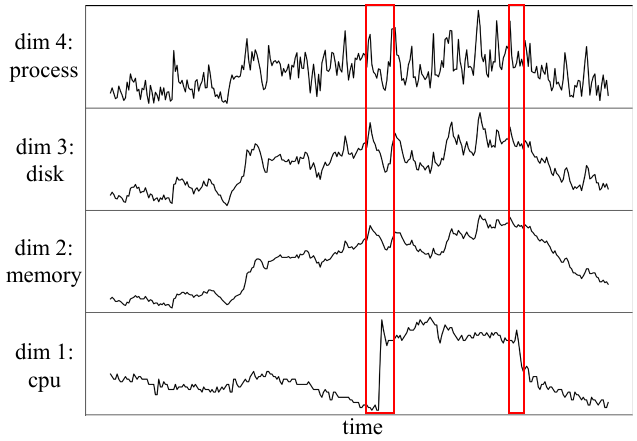}
    \caption{An MTS example from server monitor, with anomalies in red}
    \label{fig:anomaly example}
\end{figure}

A time-series is an ordered sequence with timestamped data values, widely presenting across real-world applications. With the growing demand for time-series analysis \cite{ComprehensiveSurvey2022, onlinemobile2016tong}, anomaly detection has become increasingly important in domains such as service monitoring, Internet of Things (IoT), and aerospace. Time-series anomalies, representing abnormal or unexpected behaviors, require special attentions, as exampled in \autoref{fig:anomaly example}.

Most MTS anomaly detection methods rely on a single model that extracts temporal and dimensional features through well-designed architectures, such as prediction, reconstruction, or contrastive learning-based methods.
However, no single universal model can achieve optimal performance across all types of time-series and anomaly patterns \cite{ComprehensiveSurvey2022}.
Therefore, \textit{model selection} and \textit{model ensembling} methods \cite{CAEEnsemble2021, UMSTSAD2023, AutoTSAD2024} leverage multiple models and apply selection or ensembling strategies to maintain robust anomaly detection performance, even when the data distribution significantly differs from the original training dataset.

However, existing selection and ensembling methods still face some limitations:
(i) selection methods output only a single model, making them highly sensitive to the selection strategy;
(ii) ensembling methods either combine all candidate models \cite{CAEEnsemble2021} or are restricted to univariate time-series \cite{AutoTSAD2024};
(iii) due to the constraints of candidate models, existing methods remain tied to the dimensionality of the training dataset, thereby limiting scalability.

To address the challenges above, we propose \textbf{DMPEAD}, a \textbf{D}ynamic \textbf{M}odel \textbf{P}ool and \textbf{E}nsembling framework for MTS \textbf{A}nomaly \textbf{D}etection. The framework contains three steps: 
(i) Model pool construction sequentially trains multiple basic models on different datasets from varying domains, using parameter transfer strategy and diversity metric.
(ii) Model pool updating takes new MTS as input and employs a meta-model to decide whether to expand the pool and select a suitable model subset. Meanwhile, a similarity-based strategy triggers model pool merging to preserve diversity.
(iii) Model ensembling and anomaly detection uses the selected subset for scoring, proxy metric ranking, top-k aggregation, and anomaly thresholding to produce the final anomaly detection result.
The contributions are as follows: 

(1) We propose a dynamic model pool with construction and updating, where the former builds the initial pool for broad adaptability, and the latter uses meta-model based expansion and similarity-based merging to maintain adaptability and efficiency to evolving data.

(2) We introduce a model ensembling and anomaly detection strategy that ensembles only a subset of models from the pool, rather than all models, enhancing efficiency.

(3) We conduct extensive experiments on 8 real-world MTS datasets. The results show that our model achieves a trade-off among anomaly detection performance, efficiency, generalization, and parameter robustness.

\section{Related Work}
MTS anomaly detection can be classified into \textit{single-model} and \textit{multi-model} methods, with the latter includes model selection and ensembling.

\textbf{Single-model methods} include shallow and deep learning approaches. Shallow methods rely on statistical learning, data mining, outlier detection, or classical machine learning, while deep methods include prediction-, reconstruction-, contrastive learning-based and universal time-series models, such as Anomaly Transformer \cite{AnomalyTransformer2022}, DCdetector \cite{DCdetector2023}, and TimesNet \cite{TimesNet2023}.

\textbf{Model selection methods} aim to identify the optimal model from candidate model set for a given MTS input. Based on the number of models executed during inference, they fall into two categories: multi-model and single-model execution. Multi-model methods evaluate (nearly) all candidates, yielding higher accuracy but large computing cost, e.g., TSADMS \cite{UMSTSAD2023}. In contrast, single-model methods use a pre-trained selector to choose one model directly based on data characteristics, achieving relatively lower accuracy but reduced cost. 

\textbf{Model ensembling methods} combine outputs from candidate model set to enhance anomaly detection performance, e.g., CAE-Ensemble \cite{CAEEnsemble2021} and AutoTSAD \cite{AutoTSAD2024}. The former averages anomaly scores across all models, while the latter ranks and ensembles selected models to avoid full aggregation. Our approach similarly employs ranking and ensembling but adopts a dynamic, dimension-independent model set, reducing complexity and improving scalability.

\section{Problem Definition}
A time-series $X=\left\langle X_1,X_2,\ldots,X_m \right\rangle$ refers to a sequence of data points with timestamps. In MTS, each data point $X_i$ is an $n$-dimensional vector. 
Time-series anomalies refer to data points $X_i\in X$ or subsequences $X_{i:j}\in X$ that deviate significantly from the normal distribution of $X$.

\textbf{MTS anomaly detection method} $M$ (with parameter $\theta$) takes $X$ as input, outputs either an anomaly score $S=\left\langle S_1,S_2,\ldots,S_m \right\rangle$ or a prediction label $Y=\left\langle y_1,y_2,\ldots,y_m \right\rangle$, where $y_i\in\{0,1\}$ indicates whether point $X_i$ is anomalous ($1$) or normal ($0$). Converting $S$ to $Y$ involves threshold selection and anomaly identification. Once the threshold $\varepsilon$ is set, identification is performed as follows: 
\begin{equation}
    \label{eq:anomaly thresholding}
    y_i = 
    \begin{cases}
    1 & \text{if } S_i > \varepsilon, \\
    0 & \text{else}.
    \end{cases}
\end{equation}

\textbf{MTS anomaly detection model ensembling}. Given a time-series $X$ and a set of anomaly detection methods $\mathit{MSet} = \{ M_1, M_2, \ldots, M_m \}$, model ensembling refers to $\mathit{MSetEns} = \sum{W_i M_i(X)}$, where $M_i(X)$ and $W_i$ denote the anomaly detection result and corresponding weight for the method $M_i$, and $\sum W_i$ determines the ensembling strategy.

\section{Our Method: DMPEAD}
\begin{figure}[t]
    \centering
    \includegraphics[width=0.65\textwidth]{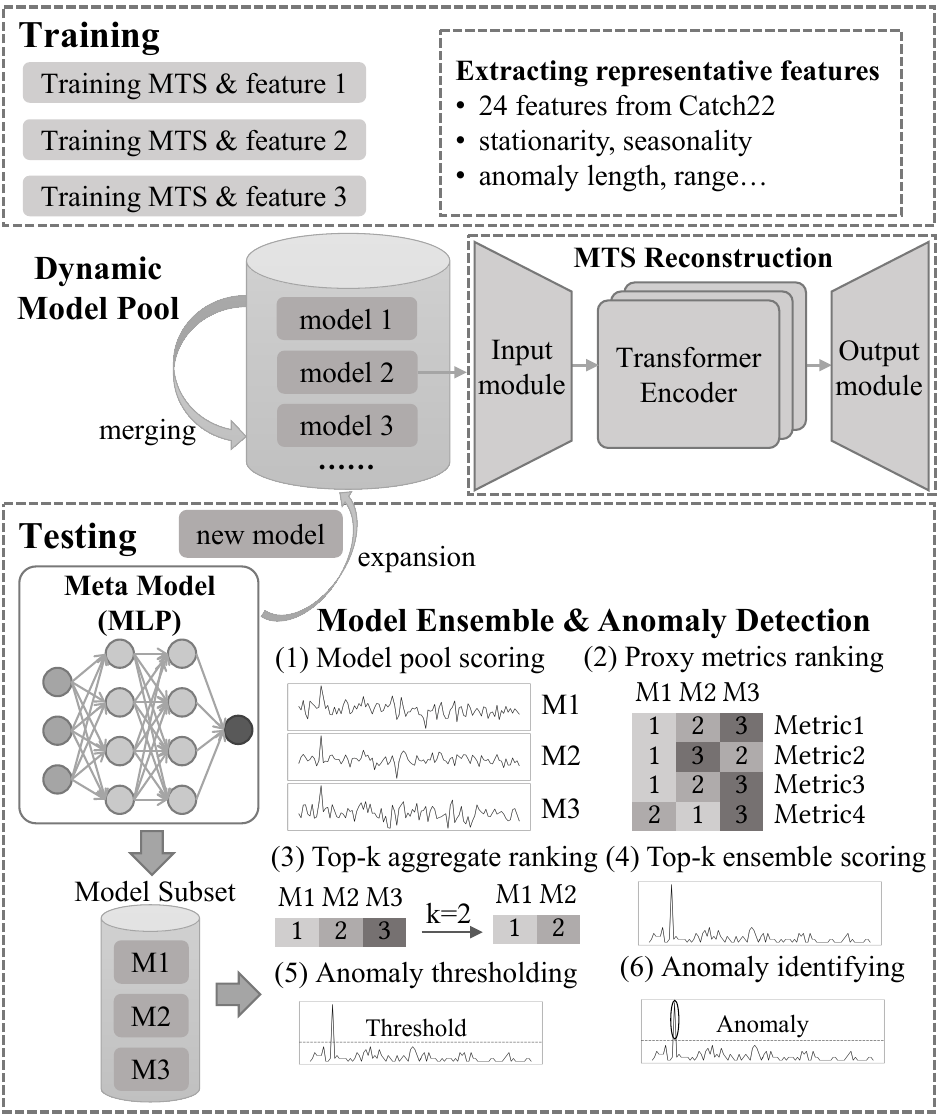}
    \caption{The proposed framework DMPEAD}
    \label{fig:framework}
\end{figure}

The proposed framework DMPEAD, as shown in \autoref{fig:framework}, consists of the following three steps: (i) model pool training and construction, (ii) model pool testing and updating, and (iii) model ensembling and anomaly detection.

\subsection{Model Pool Training and Construction}
For the initial model pool construction, we prepare multiple training dataset, use the basic model to train them separately with parameter transfer strategy and diversity metric.

\subsubsection{Dimension-independent Reconstruction Basic Model}
We choose the dimension-independent, Transformer-based reconstruction model \cite{DSDEPaper} as the basic model for the model pool and model ensembling, eliminating the reliance on the fixed data dimensionality, thereby achieving high scalability. The model comprises three parts: dimension-independent input module, Transformer encoder-based backbone, and output module, as shown in \autoref{fig:framework}.

The basic model built on dimension independence and Transformer encoder, reconstructs each MTS dimension independently, eliminating the reliance on the fixed dimensionality. Dimension independence enables the possibility of the dynamic model pool mechanism, which requires the model pool to support datasets with different dimensions. Additionally, The model also transforms original point-wise inputs into segment-wise ones for each MTS dimension, reducing both time and space complexity.

During the training, MSE reconstruction loss is applied as the training loss, where $\hat{X}$ is the reconstructed data.
\begin{equation}
    Loss_{\mathrm{MSE}}=
    \mathrm{MSE}\left(X,\hat{X}\right)
    =\sum_{i=1}^{m}\mathrm{MSE}\left(X_i,{\hat{X}}_{i}\right)
    \label{eq:MSE reconstruction loss}
\end{equation}

\subsubsection{Initial Model Pool Construction}
We sequentially train the basic model on multiple training MTS to obtain corresponding trained models, which be parts of initial model pool. To accelerate the training process, each model is trained based on the previous model via \textit{parameter transfer strategy}. Additionally, to ensure model diversity, the training loss includes not only the reconstruction loss between input and reconstructed data, but also a \textit{diversity metric} between the current training model and previous trained models in the pool. 

\textbf{Parameter transfer strategy}. Given a trained model $M_{i-1}$ (parameters $\theta_{i-1}$), model $M_i$ with parameters $\theta_{i}$ receives a randomly selected factor $\beta$ of parameters from $\theta_{i-1}$, which are then frozen. During the $M_i$ training, only the remaining $1 - \beta$ factor of parameters are updated iteratively. This parameter transfer strategy significantly improves multi-model training efficiency compared to purely sequential training, where models are trained without parameter sharing.
Differing from \cite{CAEEnsemble2021}, we use various datasets to train the basic model, generate trained models that be applicable to the corresponding dataset and domain. 

\textbf{Diversity metric}. Inspired by \cite{CAEEnsemble2021}, a model ensembling method benefits from consisting of diverse basic models in the pool. The more diverse the basic models in the pool are, the higher the accuracy that model ensembling can achieve. In MTS anomaly detection, model diversity can be reflected as the anomaly score and reconstruction error. We define diversity metric $Div_{M_i, M_j}$ to measure the dissimilarity between two basic models $M_i$ and $M_j$:
\begin{equation}
    Div_{M_i, M_j}(X) = \mathrm{MSE}\left({M_i}(X), {M_j}(X)\right)
    \label{eq:Diversity metric of tow models}
\end{equation}

where ${M_i}(X)$ and ${M_j}(X)$ denote the reconstructed data of $M_i$ and $M_j$ on $X$. The larger $Div_{M_i, M_j} (X)$ means the larger difference between $M_i$ and $M_j$.

\textbf{Model pool training loss}. Combined the reconstruction loss \autoref{eq:MSE reconstruction loss} and diversity metric \autoref{eq:Diversity metric of tow models}, we can obtain the model pool training loss, where $\mu$ indicates the importance of the two components:
\begin{equation}
    \mathcal{L}_i = Loss_{\mathrm{MSE}} - \mu\sum_{j=1}^{i-1}{Div_{M_i, M_j}(X)}
    \label{eq:Model pool training loss}
\end{equation}

\subsection{Model Pool Testing and Updating}
The automatic update mechanism dynamically adjusts the model pool during testing through (i) \textbf{model pool expansion}, adding new model, and (ii) \textbf{model pool merging}, reducing similar models, ensuring reliability and efficiency under distribution shifts in incoming MTS. Notably, pool expansion does not retrain or modify existing models.

\subsubsection{Model Pool Expansion}
Model pool expansion involves decision-making for new model creation, and the creating and training strategy.

\textbf{Triggering new model creation} can initially be made by directly evaluating the output of models in the pool, e.g., reconstructed data or anomaly score. However, this approach proves inefficient and time cost as it need to execute all models \cite{AutoTSAD2024}. Thus, we propose a \textit{meta-model} that leverages dataset characteristic and model fingerprint, requiring executing only a subset of models. Our solution involves: 
(i) constructing meta-model training data;
(ii) training MLP-based meta-model;
(iii) performing meta-model to select suitable models and make decision for new model creation;
(iv) synchronous updating meta-model and its training data.

Constructing meta-model training data includes extracting representative features for each dataset and generating model fingerprints. The representative features involves multiple features from Catch22 \cite{Catch222019}, stationary, seasonal, and anomaly information, including cross-dimension and inner-dimension features. Model fingerprint indicates the model behavior and feature, which is generated by a standard valid MTS, outputting as the representative features of the MTS reconstruction error.

The MLP-based meta-model is trained on above training data. To ensure reliability and prevent overfitting, we employ standard strategies: K-fold cross-validation (K=5 by default), learning rate scheduling, and early stopping with patience. The final model is selected from the fold yielding the lowest validation loss.

We perform meta-model to quantify the adaptability of basic models to new MTS. After being trained, the meta-model accepts new MTS as input and outputs the predicted error for each basic model. Then, a basic model is considered compatible with new MTS, and consequently added to the matched model subset $\mathit{MSet}'$, if the negation of its predicted error (e.g., model match score $\mathit{MS}(M_i)$) is above the model-matching threshold $\varepsilon_{\mathrm{model}}$, i.e., $\mathit{MS}(M_i)>\varepsilon_{\mathrm{model}}$.
Furthermore, a decision is made on whether to trigger the new model creation: if the size of the subset $\mathit{MSet}'$ is above the matching-count threshold $\varepsilon_{\mathrm{judge}}$, i.e., $\left|\mathit{MSet}'\right| > \varepsilon_{\mathrm{judge}}$, the pool is deemed sufficiently adaptive to new MTS, and no new model is created; otherwise, a new model is created and trained to maintain the adaptability of the pool.

Considering the dynamic nature of the model pool, the meta-model must also remain adaptive. When new MTS necessitates the new model creation, the corresponding data features and model fingerprints are incorporated into the meta-model training data, triggering the meta-model update to accommodate the changes introduced by new MTS.

\textbf{New model creation and training}. We adopt the same structure \cite{DSDEPaper} as the basic model. The training process follows the initial model pool construction, incorporating parameter transfer and diversity metric, which helps preserve the dynamic nature of the pool. Notably, our parameter transfer strategy can either transfer parameters from the last model, or average weights across all models in the pool.

\subsubsection{Model Pool Merging}

Model pool merging involves merging similar models and synchronously updating the meta-model.

\textbf{Similar model merging} consists of four steps: (i) determine whether to trigger model merging; (ii) compute pairwise model dissimilarity based on model parameters; (iii) decide whether to merge using a threshold-based strategy; and (iv) execute the merging and update the model pool and associated components.

Model merging can be triggered by a simple threshold strategy: if the model pool size exceeds the model-merging threshold ${\varepsilon_{\mathrm{merge}}}$, i.e., $\left|\mathit{MSet}\right| > {\varepsilon_{\mathrm{merge}}}$, merging is initiated to reduce redundancy and computational overhead. To maintain the model pool consistency during one MTS test, merging is performed either before or after the test phase.

When merging is triggered, pairwise model dissimilarity is computed based on model parameters $(\theta_i, \theta_j)$ for each two models $(M_i, M_j)$ in the pool, comprising three components: (i) Euclidean distance, (ii) statistical differences, e.g., mean, standard deviation, min/max, and (iii) cosine dissimilarity, i.e., $(1 - \mathit{cosine}(\theta_i, \theta_j))$. Each component is normalized, and their average yields the dissimilarity score $DS(M_i, M_j)$ for model pair $(M_i, M_j)$.

After computing the dissimilarity scores, a threshold-based strategy determines which model pairs to merge: if the score belows the dissimilarity-score threshold $\varepsilon_{\mathrm{disscore}}$, i.e., $DS(M_i, M_j) < \varepsilon_{\mathrm{disscore}}$, models $M_i$ and $M_j$ are deemed sufficiently similar and merged. To avoid aggressive compression (chaining merges), each model is allowed to participate in at most one merge per round. Consequently, the number of models reduced in a single merging is at most half the pool size.

For the model pair $(M_i, M_j)$ selected for merging, their parameters are averaged to form the new model $M_{\mathrm{new}}$, i.e., $\theta_{\mathrm{new}} = (\theta_i + \theta_j)/2$. After merging, $M_i$ and $M_j$ are removed from the pool and replaced by $M_{\mathrm{new}}$. Notably, since merging may occur multiple rounds, $M_i$ and $M_j$ could be merged models from prior rounds, indicating higher similarity.

\textbf{Synchronous meta-model update}. Following pool merging, the representative dataset features in the meta-model training data remain unchanged, preserving the incremental state from new model creation and retaining broad insight into model adaptability. However, the model pool must be updated to reflect the latest model fingerprints, triggering a complete refresh of the meta-model training data. This update is costlier than the one triggered by pool expansion and necessitates retraining the meta-model.

\subsection{Model Ensembling and Anomaly Detection}
We employ model ensembling within the model pool for anomaly detection. Unlike existing methods \cite{CAEEnsemble2021} that aggregate scores from all basic models, potentially including weak performers, we ensemble only the top-k ranked models from the selected subset. As anomaly labels are typically unavailable, models are ranked in an unsupervised manner using multiple proxy metrics to enhance reliability.
The top-k model ensembling and anomaly detection process includes (\autoref{fig:framework}):
(i) obtaining anomaly scores from models in the subset;
(ii) computing multiple proxy metrics to produce independent rankings;
(iii) applying robust rank aggregation to combine rankings, select the top-k models, and average their scores to obtain the final anomaly score;
(iv) determining an anomaly threshold from the aggregated score and identifying anomalies.

\textbf{Model pool scoring} computes anomaly score for each model in the subset. The anomaly score quantifies the anomaly degree of each data point, which higher values indicates higher anomaly likelihood. MSE is used for score calculation, consistent with the model reconstruction loss.

\textbf{Proxy metrics ranking} calculates three types of proxy metrics to generate several rankings.
\textit{Prediction error} measures the discrepancy between input and reconstructed data using 4 metrics: MSE, MAE, RMSE, and MAPE.
\textit{Synthetic anomaly detection} evaluates model performance on synthetic data with 5 anomaly types: spikes, contextual, flip, speedup, and scale.
\textit{Model centrality} reflects the relative position of models within a cluster, which “good” models tend to cluster near the center, while “bad” ones are more dispersed. We measure centrality by clustering, using 4 methods: nearest neighbor, k-medoids, affinity propagation, and greedy farthest distance clustering.

\textbf{Top-k rank aggregation} combines proxy metric rankings through robust aggregation to select the top-k models and averages their anomaly scores to produce the final result. We adopt the Top-k Borda Rank method, which outperforms other aggregation methods \cite{UMSTSAD2023}. If $k$ exceeds the size of the subset $\left|\mathit{MSet}'\right|$, ranking and aggregation are skipped, and direct averaging is applied.

\textbf{Anomaly thresholding and identification} selects the anomaly threshold from the aggregated scores and identifies anomalies accordingly.
\textit{Threshold selection} uses three methods: (i) mean-standard deviation method sets the threshold as the mean plus 2.5 times standard deviation; (ii) epsilon method searches for a threshold balancing point and sequence anomalies; and (iii) percentile method defines the threshold by a specified anomaly ratio.
\textit{Anomaly identification} labels data points with scores exceeding the threshold as anomalies, as shown in \autoref{eq:anomaly thresholding}.

\section{Experiments}

\begin{table}[htbp]
\centering
\caption{MTS anomaly detection dataset statistics}
\label{tab:dataset_info}
\setlength{\tabcolsep}{5.5pt}
\begin{tabular}{lcccccc}
\hline
Dataset     & Domain        & Dim.  & MTS Num.  & Avg. Len. & AR(\%) \\
\hline
Exathlon    & monitor       & 19    & 31        & 25,479    & 8.07 \\
PSM         & monitor       & 25    & 1         & 220,322   & 13.87 \\
SMD         & monitor       & 38    & 28        & 25,300    & 2.10 \\
SMAP        & aerospace     & 25    & 54        & 5,313     & 6.20 \\
MSL         & aerospace     & 55    & 27        & 2,445     & 6.01 \\
SWaT        & IoT           & 51    & 1         & 944,919   & 6.07 \\
WADI        & IoT           & 123   & 1         & 957,372   & 2.88 \\
UAV         & aircraft      & 33-41 & 13        & 99,400    & 34.85 \\
\hline
\end{tabular}
\end{table}

We evaluate DMPEAD on 8 real-world MTS anomaly detection datasets from diverse domains, including service monitor, aerospace, IoT, and aircraft, as shown in \autoref{tab:dataset_info}, where UAV is our collected dataset. 
We select 3 metrics for extensive evaluation: TS-AUC-PR, Range-AUC-PR, and VUS-PR \cite{VUS2022}.
We compare our model with 6 baseline methods: Anomaly Transformer \cite{AnomalyTransformer2022}, DCdetector \cite{DCdetector2023}, TimesNet \cite{TimesNet2023}, Patchformer \cite{DSDEPaper}, CAE-Ensemble \cite{CAEEnsemble2021}, and TSADMS \cite{UMSTSAD2023}.

\textbf{Comparison experiment}. Due to the reliance of baselines on dimensionality as fixed parameter, and the mismatch in dimensions between the UAV train and test MTS, the experiment uses the first 7 datasets. Specifically, performance of our model and 6 baselines is evaluated across 143 test MTS from 7 datasets, as shown in \autoref{tab:comparison result}. 
The bold values denote the best performance for each dataset and metric. 
For TS-AUC-PR, our model achieves the best on Exathlon, SWaT, and WADI. For Range-AUC-PR, it performs best on PSM, SMD, and WADI. For VUS-PR, our model leads on PSM and WADI. 
Overall, our model obtains 8 best across all datasets and metrics, outperforming all baselines, highlighting strong performance.

\begin{table}[htbp]
\centering
\caption{Performance comparison using 7 datasets and 3 metrics TS-AUC-PR(TS), Range-AUC-PR(R), VUS-PR(V)}
\label{tab:comparison result}
\setlength{\tabcolsep}{1pt}
\renewcommand{\arraystretch}{1}
\begin{tabular}{lccccccccccccccccccccc}
\hline
Dataset & \multicolumn{3}{c}{A. Trans.\cite{AnomalyTransformer2022}} & \multicolumn{3}{c}{DCdetect.\cite{DCdetector2023}} & \multicolumn{3}{c}{TimesNet\cite{TimesNet2023}} & \multicolumn{3}{c}{Patchfor.\cite{DSDEPaper}} & \multicolumn{3}{c}{CAE-Ens.\cite{CAEEnsemble2021}} & \multicolumn{3}{c}{TSADAMS\cite{UMSTSAD2023}} & \multicolumn{3}{c}{\textbf{Ours}} \\
\hline
 & TS & R & V & TS & R & V & TS & R & V & TS & R & V & TS & R & V & TS & R & V & TS & R & V \\
\hline
Exathlon & 0.52 & 0.27 & 0.27 & 0.58 & 0.18 & 0.16 & 0.56 & 0.61 & 0.63 & 0.55 & 0.61 & 0.63 & 0.61 & \textbf{0.64} & \textbf{0.66} & 0.50 & 0.33 & 0.34 & \textbf{0.61} & 0.60 & 0.61 \\
MSL & 0.53 & 0.22 & 0.19 & 0.56 & 0.19 & 0.15 & \textbf{0.58} & 0.40 & 0.37 & 0.58 & \textbf{0.40} & \textbf{0.38} & 0.56 & 0.28 & 0.24 & 0.43 & 0.31 & 0.30 & 0.56 & 0.35 & 0.30 \\
PSM & 0.64 & 0.40 & 0.37 & 0.26 & 0.37 & 0.32 & 0.64 & 0.55 & 0.54 & 0.64 & 0.49 & 0.48 & 0.64 & 0.42 & 0.39 & \textbf{0.64} & 0.53 & 0.53 & 0.64 & \textbf{0.57} & \textbf{0.56} \\
SMAP & 0.55 & 0.15 & 0.15 & 0.56 & 0.18 & 0.18 & \textbf{0.58} & \textbf{0.26} & \textbf{0.24} & 0.58 & 0.24 & 0.23 & 0.56 & 0.20 & 0.19 & 0.35 & 0.21 & 0.19 & 0.56 & 0.24 & 0.22 \\
SMD & 0.44 & 0.10 & 0.08 & 0.29 & 0.07 & 0.05 & 0.49 & 0.49 & \textbf{0.45} & 0.50 & 0.48 & 0.45 & \textbf{0.52} & 0.29 & 0.25 & 0.47 & 0.32 & 0.31 & 0.52 & \textbf{0.50} & 0.44 \\
SWaT & 0.56 & 0.21 & 0.22 & 0.56 & 0.31 & 0.33 & 0.56 & 0.11 & 0.10 & 0.56 & 0.10 & 0.09 & 0.56 & \textbf{0.38} & \textbf{0.42} & 0.06 & 0.10 & 0.08 & \textbf{0.56} & 0.10 & 0.09 \\
WADI & 0.14 & 0.08 & 0.08 & 0.53 & 0.07 & 0.06 & 0.27 & 0.24 & 0.25 & 0.08 & 0.19 & 0.19 & 0.53 & 0.13 & 0.14 & 0.03 & 0.07 & 0.06 & \textbf{0.53} & \textbf{0.25} & \textbf{0.25} \\
\hline
\end{tabular}
\end{table}

\begin{table}[htbp]
\centering
\caption{Total runtime during train and test phases}
\label{tab:runtime result}
\setlength{\tabcolsep}{8pt}
\begin{tabular}{lcc}
\hline
Model & Train runtime (s) & Test runtime (s) \\
\hline
AnomalyTrans. \cite{AnomalyTransformer2022} & \textbf{533} & \underline{377} \\
DCdetector \cite{DCdetector2023} & 49978 & 1147 \\
TimesNet \cite{TimesNet2023} & 11627 & 2205 \\
Patchformer \cite{DSDEPaper} & 3356 & \textbf{343} \\
CAEEnsemble \cite{CAEEnsemble2021} & 2691 & 609 \\
TSADAMS \cite{UMSTSAD2023} & 3811 & 15895 \\
\textbf{DMPEAD (Ours)} & \underline{1362} & 2146 \\
\hline
\end{tabular}
\end{table}

We also report the total train and test runtime in \autoref{tab:runtime result}. Our model, trained once via parameter transfer, ranks second in training efficiency. In testing, although slightly slow, it outperforms the worst two baselines, mainly due to the meta-model and proxy-metric ranking steps.

\textbf{Ablation experiment}. To clarify the individual contribution of each component to performance, we conduct ablation studies on its main modules. The experiment is organized on module-level items:  
(i) Disabling model pool, i.e., \textit{w/o pool}, reducing to a single model (i.e., Patchformer \cite{DSDEPaper}), also disabling both pool update mechanism and ensembling;
(ii) Disabling pool expansion, i.e., \textit{w/o expansion}, also disabling pool merging, as the initial pool typically requires no merging;  
(iii) Disabling pool merging, i.e., \textit{w/o merging}.

\autoref{tab:ablation result} reports performance and runtime comparisons of ablations across all datasets under TS-AUC-PR and Range-AUC-PR. Metric results of ablation items are presented as the relative difference compared to non-ablated item. Bold values denote notable performance degradation or runtime increases after ablation.
\textit{w/o pool} causes the largest performance loss, reaching 7.7\% on PSM (Range-AUC-PR) and 44\% on WADI (TS-AUC-PR), and largely increases train runtime due to the loss of accelerated training.
\textit{w/o expansion} leads to a moderate decline, around 3\% on PSM under Range-AUC-PR.
\textit{w/o merging} has little effect on accuracy but significantly increases test runtime, indicating that pool merging mainly improves efficiency by reducing redundancy, with minimal effect on accuracy.
Overall, the model pool and its update mechanism are both critical for balancing performance and efficiency.

\begin{table}[htbp]
\centering
\caption{Ablation result on TS-AUC-PR (TS) and Range-AUC-PR (R) metrics}
\label{tab:ablation result}
\setlength{\tabcolsep}{7pt}
\renewcommand{\arraystretch}{1}
\begin{tabular}{lcccccccc}
\hline
Ablation & \multicolumn{2}{c}{Non} & \multicolumn{2}{c}{\textit{w/o pool}} & \multicolumn{2}{c}{\textit{w/o expansion}} & \multicolumn{2}{c}{\textit{w/o merging}} \\
\hline
 & TS & R & TS & R & TS & R & TS & R  \\
\hline
Exathlon & 0.61 & 0.60 & \textbf{-0.060} & +0.007 & 0.000 & +0.004 & 0.000 & 0.000 \\
MSL & 0.56 & 0.35 & +0.019 & +0.053 & 0.000 & +0.004 & 0.000 & 0.000 \\
PSM & 0.64 & 0.57 & 0.000 & \textbf{-0.077} & 0.000 & \textbf{-0.032} & 0.000 & 0.000 \\
SMAP & 0.56 & 0.24 & +0.013 & +0.002 & 0.000 & +0.001 & 0.000 & 0.000 \\
SMD & 0.52 & 0.50 & \textbf{-0.020} & \textbf{-0.021} & -0.001 & -0.003 & 0.000 & -0.004 \\
SWaT & 0.56 & 0.10 & 0.000 & -0.001 & 0.000 & 0.000 & 0.000 & 0.000 \\
UAV & 0.71 & 0.42 & 0.000 & -0.001 & 0.000 & -0.003 & 0.000 & -0.001 \\
WADI & 0.53 & 0.25 & \textbf{-0.444} & \textbf{-0.060} & 0.000 & -0.004 & 0.000 & +0.002 \\
Train time (s) & \multicolumn{2}{c}{1362} & \multicolumn{2}{c}{\textbf{3356}} & \multicolumn{2}{c}{1362} & \multicolumn{2}{c}{1362} \\
Test time (s) & \multicolumn{2}{c}{2146} & \multicolumn{2}{c}{343} & \multicolumn{2}{c}{2295} & \multicolumn{2}{c}{\textbf{2753}} \\
\hline
\end{tabular}
\end{table}

\begin{figure}[htbp]
    \centering
    \begin{subfigure}[b]{0.48\textwidth}
        \centering
        \includegraphics[width=\linewidth]{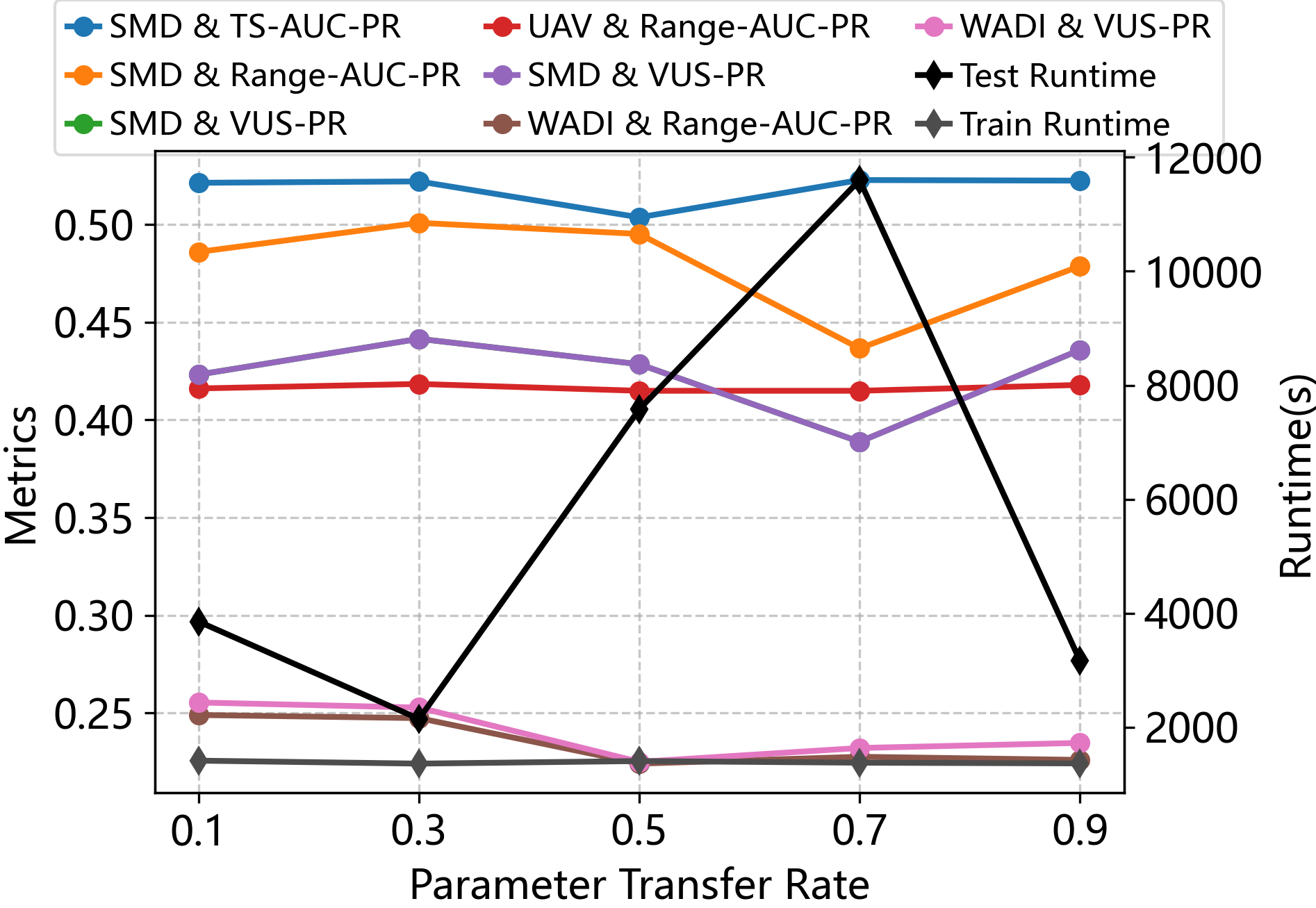}
        \caption{Varying $\beta$, fix $\mu$}
    \end{subfigure}
    \hfill
    \begin{subfigure}[b]{0.48\textwidth}
        \centering
        \includegraphics[width=\linewidth]{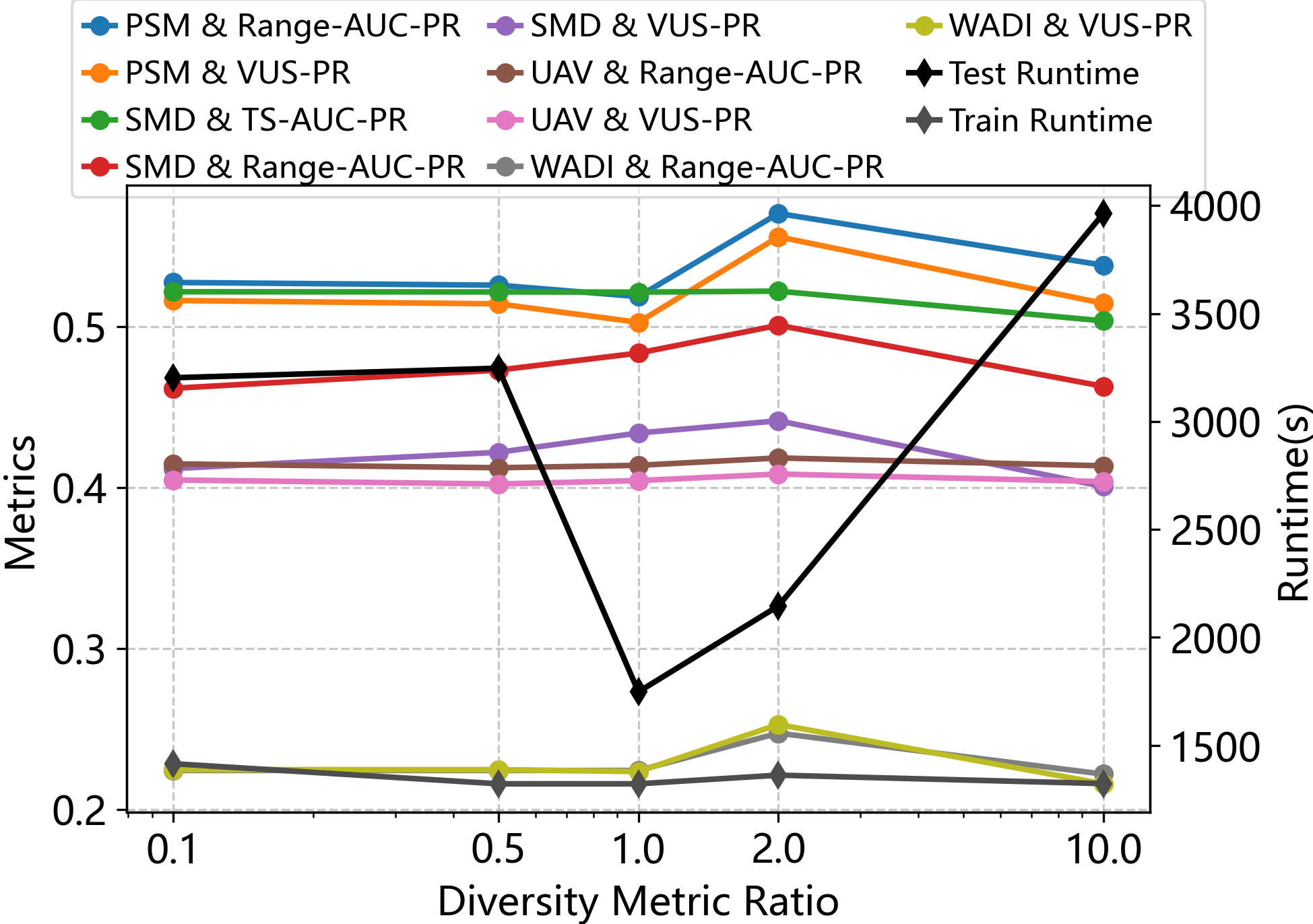}
        \caption{Varying $\mu$, fix $\beta$}
    \end{subfigure}
    \caption{Parameter sensitivity result on pool construction}
    \label{fig:sensitivity pool construction}
\end{figure}

\textbf{Parameter sensitivity experiment}. We examine 3 groups of hyper-parameters governing our model to assess performance and efficiency impact: 
(i) pool construction: parameter transfer rate $\beta$ and diversity metric ratio $\mu$;
(ii) pool expansion: model-matching threshold $\varepsilon_{\mathrm{model}}$ and matching-count threshold $\varepsilon_{\mathrm{judge}}$;
(iii) pool merging: model-merging threshold $\varepsilon_{\mathrm{merge}}$ and dissimilarity-score threshold $\varepsilon_{\mathrm{disscore}}$.
Given the prohibitively large Cartesian product of all parameter combinations, we adopt a pairwise tuning strategy in each group, fixing one parameter while varying the other, to efficiently identify a near-optimal configuration at minimal experimental cost.  

\begin{figure}[htbp]
    \centering
    \begin{subfigure}[b]{0.48\textwidth}
        \centering
        \includegraphics[width=\linewidth]{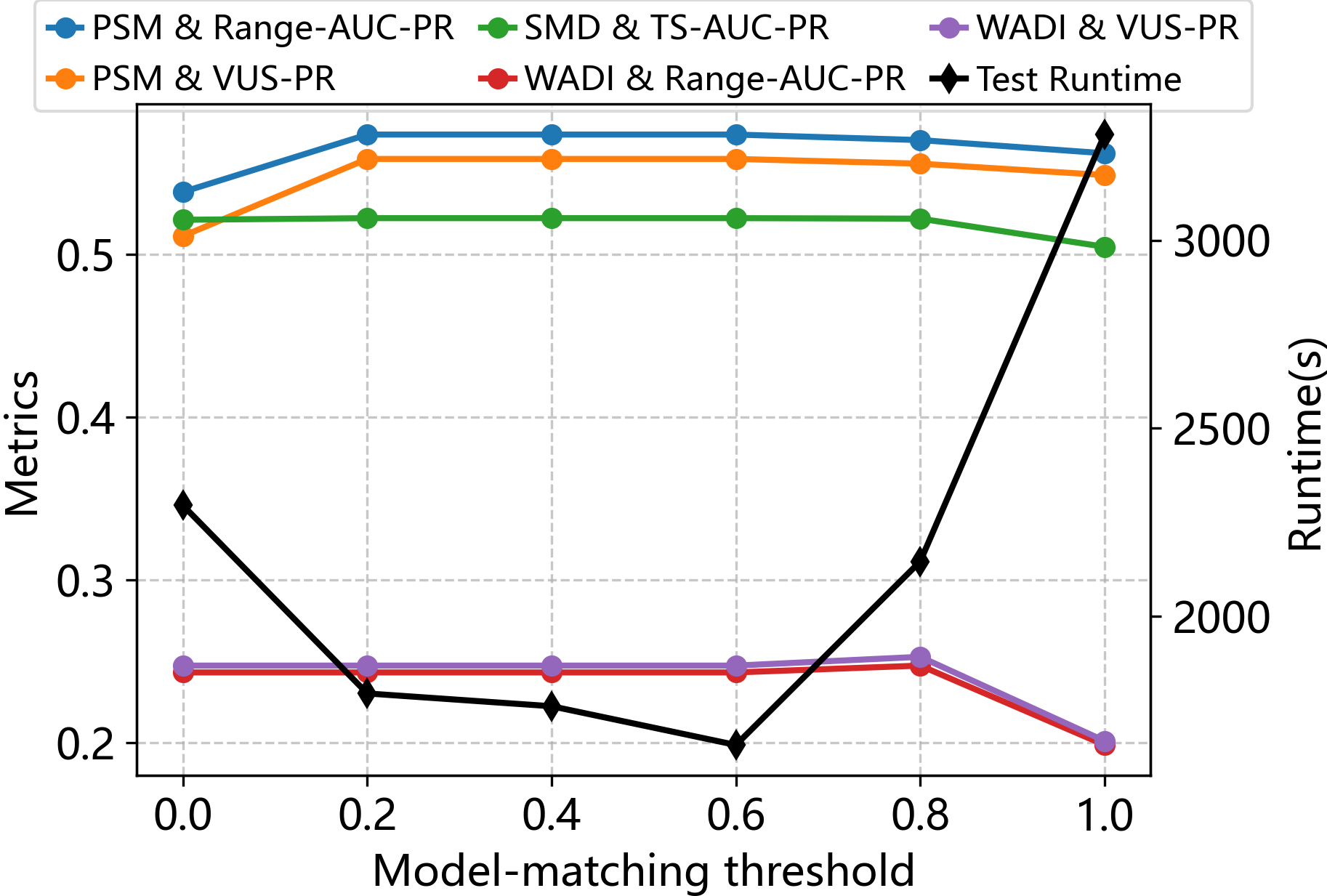}
        \caption{Varying $\varepsilon_{\mathrm{model}}$, fix $\varepsilon_{\mathrm{judge}}$}
    \end{subfigure}
    \hfill
    \begin{subfigure}[b]{0.48\textwidth}
        \centering
        \includegraphics[width=\linewidth]{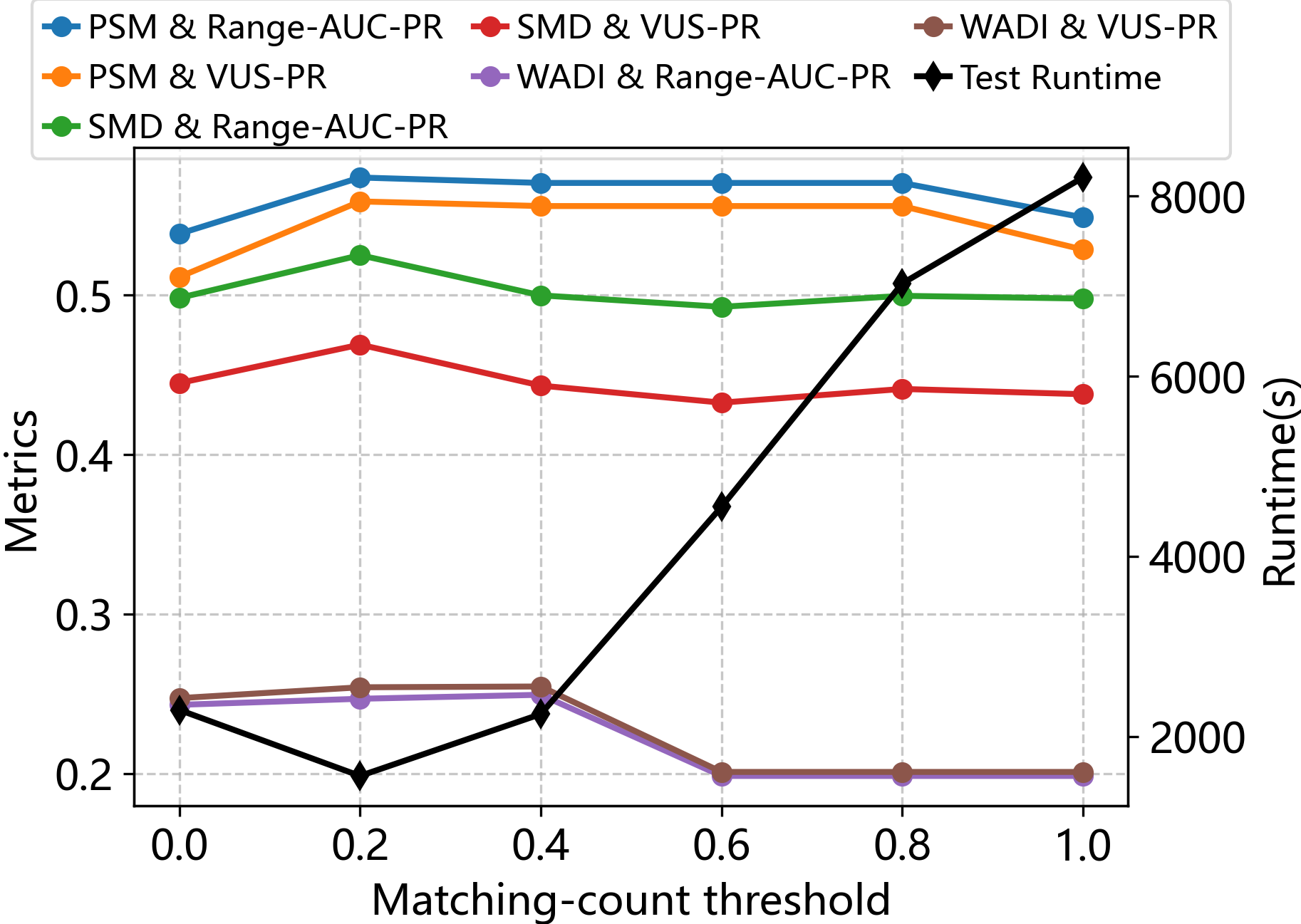}
        \caption{Varying $\varepsilon_{\mathrm{judge}}$ (requires times $\left|\mathit{MSet}\right|$), fix $\varepsilon_{\mathrm{model}}$}
    \end{subfigure}
    \caption{Parameter sensitivity result on pool expansion}
    \label{fig:sensitivity pool expansion}
\end{figure}

\begin{figure}[htbp]
    \centering
    \begin{subfigure}[b]{0.48\textwidth}
        \centering
        \includegraphics[width=\linewidth]{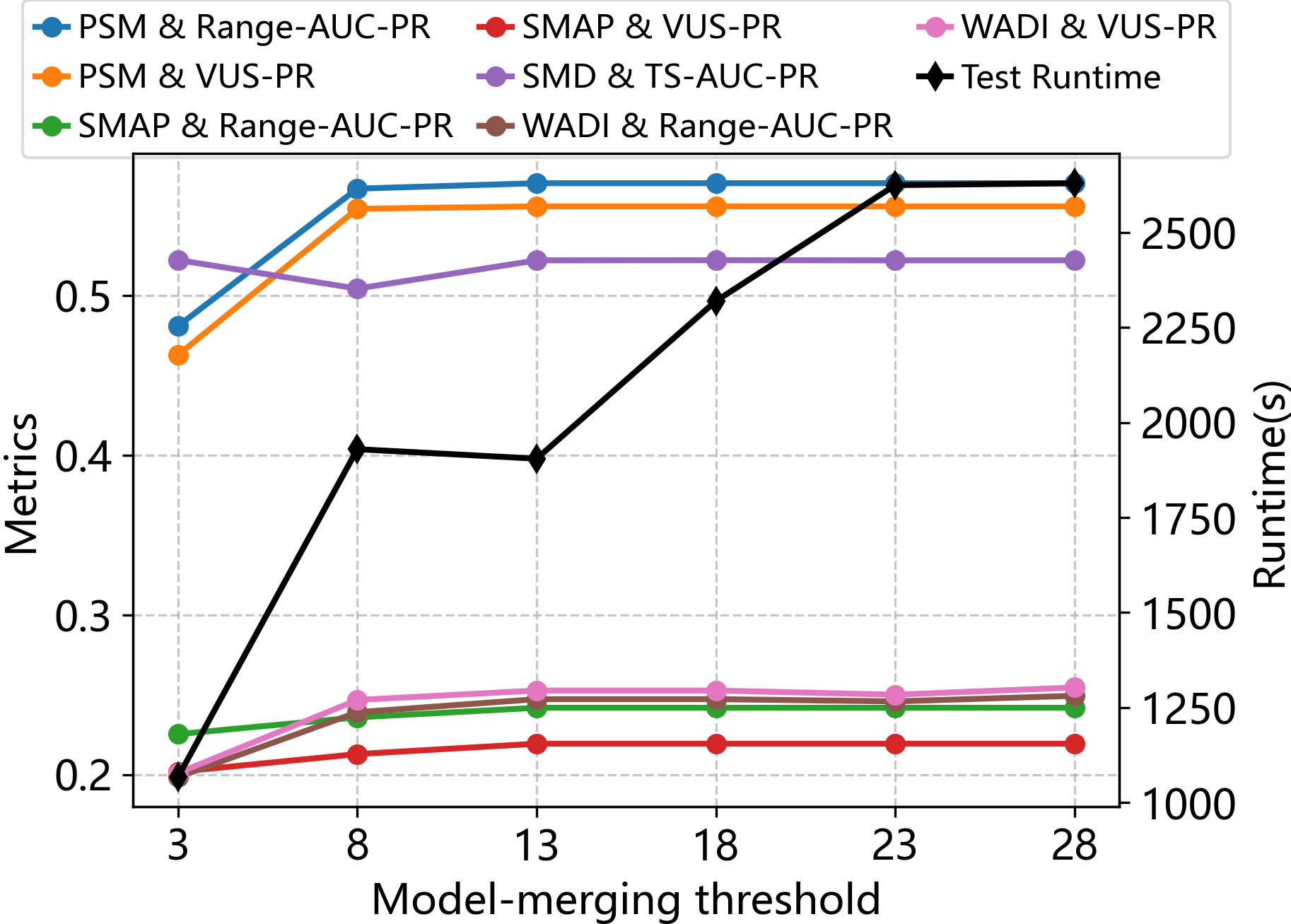}
        \caption{Varying $\varepsilon_{\mathrm{merge}}$, fix $\varepsilon_{\mathrm{disscore}}$}
    \end{subfigure}
    \hfill
    \begin{subfigure}[b]{0.48\textwidth}
        \centering
        \includegraphics[width=\linewidth]{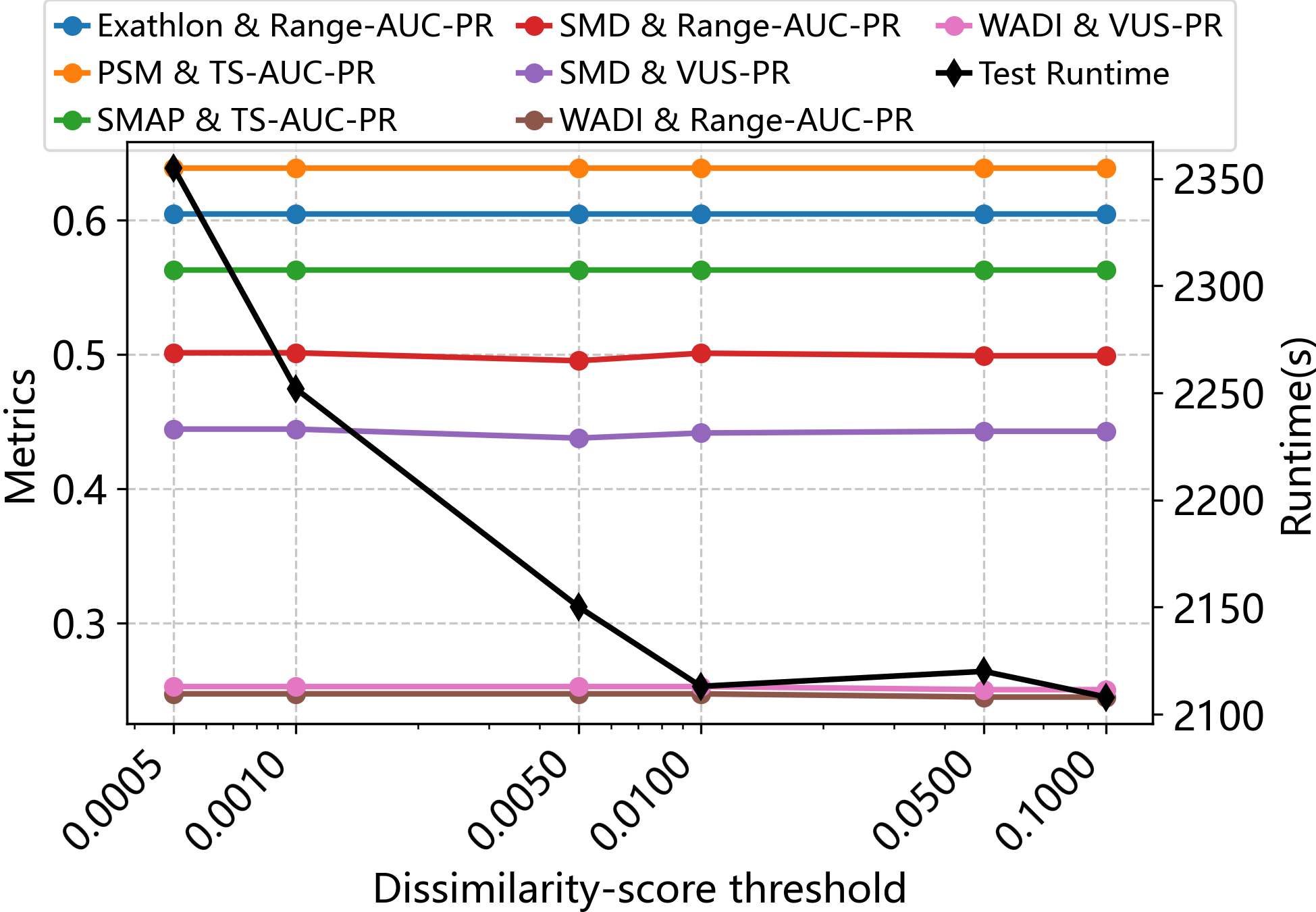}
        \caption{Varying $\varepsilon_{\mathrm{disscore}}$ (log-scale), fix $\varepsilon_{\mathrm{merge}}$}
    \end{subfigure}
    \caption{Parameter sensitivity result on pool merging}
    \label{fig:sensitivity pool merging}
\end{figure}

\autoref{fig:sensitivity pool construction} to \autoref{fig:sensitivity pool merging} illustrate selected results on several datasets and metrics, where X-axis denotes parameter values, while Y-axis shows anomaly detection metrics and total train/test runtime.
For $\beta$ and $\mu$ (\autoref{fig:sensitivity pool construction}), the optimal pool construction parameters are $(0.3, 2.0)$.
For $\varepsilon_{\mathrm{model}}$ and $\varepsilon_{\mathrm{judge}}$ (\autoref{fig:sensitivity pool expansion}), the optimal pool expansion parameters are $(0.8, 0.34*\left|\mathit{MSet}\right|)$.
For $\varepsilon_{\mathrm{merge}}$ and $\varepsilon_{\mathrm{disscore}}$ (\autoref{fig:sensitivity pool merging}), the optimal pool merging parameters are $(15, 0.01)$.

\textbf{Scalability experiment}. To assess the scalability of our model, i.e., the ability to generalize to unseen data, we conduct the leave-one-dataset-out experiment on the first 7 datasets. 
For each target dataset, the model pool is constructed on the remaining 7 datasets (including UAV) and evaluated on the held-out test MTS. Results are compared with: (i) 6 baselines trained and tested on the target dataset (shown via boxplots with mean and median), and (ii) standard DMPEAD trained with the target dataset ("DMPEAD", gray dashed line). The scalability variant ("DMPEAD Scal.", black dashed line) is evaluated solely on unseen data.

\begin{figure}[t]
    \centering
    \includegraphics[width=0.7\textwidth]{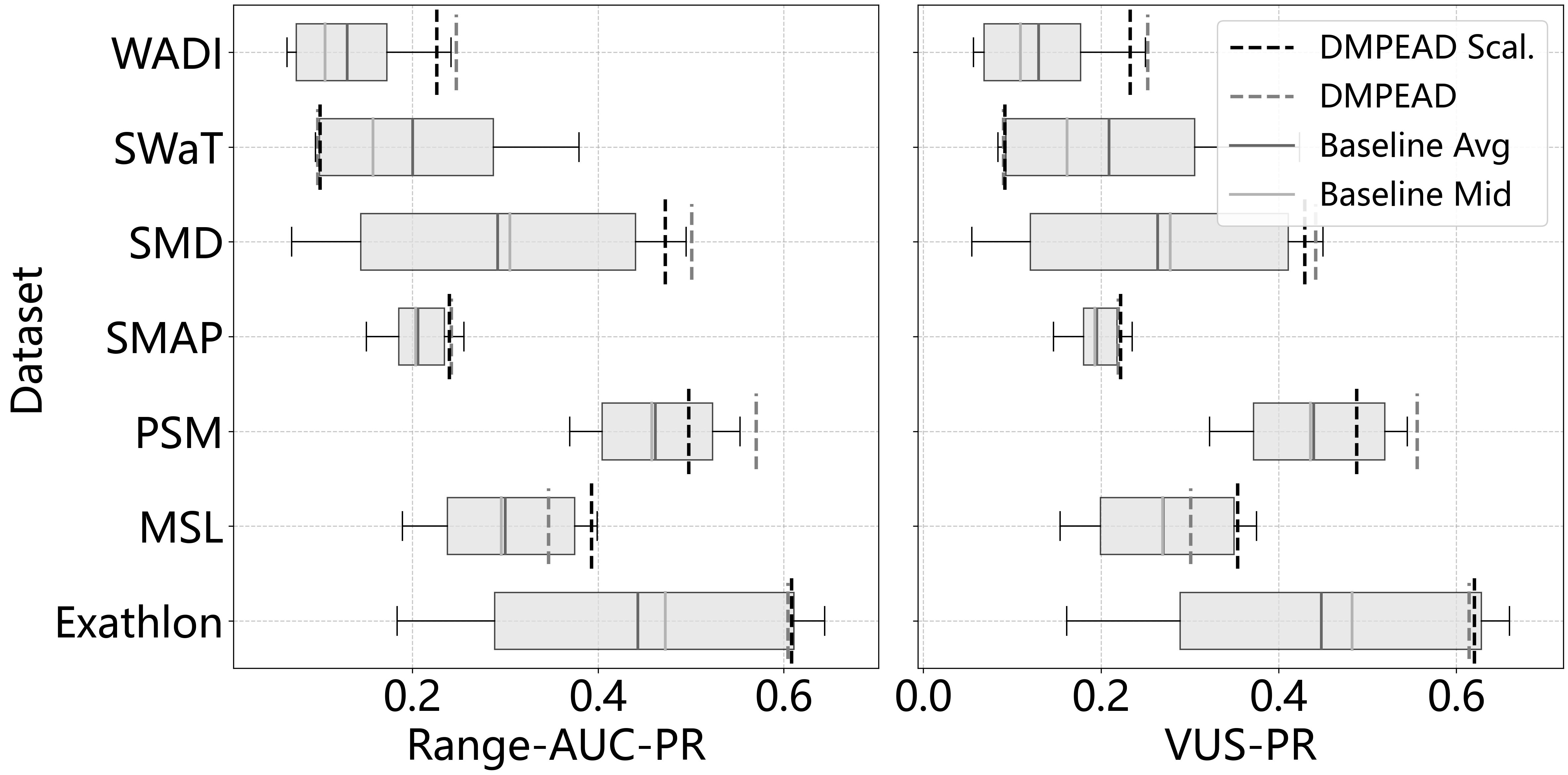}
    \caption{Scalability result on unseen data during training}
    \label{fig:scalability experiment}
\end{figure}

A black dashed line exceeds the baseline mean/median indicates that DMPEAD outperforms at least half of the baselines without in-domain training, demonstrating strong cross-dataset adaptability. When it even surpasses the gray dashed line, it highlights exceptional generalization.
\autoref{fig:scalability experiment} shows that, for Range-AUC-PR and VUS-PR, "DMPEAD Scal." consistently exceeds baseline mean/median on all datasets except SWaT, and beats "DMPEAD" on MSL and Exathlon, confirming that DMPEAD achieves robust performance on unseen data, underscoring its practical scalability.

\section{Conclusion}
We propose a dynamic model pool and ensembling framework for MTS anomaly detection, which adaptively updates and ensembles models for diverse MTS. Extensive experiments on anomaly detection comparison, ablation study, parameter sensitivity, and scalability verify its adaptability and generalization.

\bibliographystyle{splncs04}
\bibliography{reference} 

\end{document}